\documentclass[a4paper,pagenum,french]{rnti}

\usepackage{graphicx}
\usepackage{amssymb, amsthm, amsmath, amsfonts}

\usepackage[T1]{fontenc}
\usepackage[latin1]{inputenc}

\usepackage{diagbox}

\usepackage{url}

\titrecourt{MONITOR}

\nomcourt{C. Thierry et al.}

\titre{Modélisation de l'incertitude et de l'imprécision de données de crowdsourcing~: MONITOR}

\auteur{Constance Thierry\affil{1},
        Jean-Christophe Dubois\affil{1},
        Yolande Le Gall\affil{1},
         Arnaud Martin\affil{1}}

\affiliation{
    \affil{1}Université de Rennes 1, 2 rue du Thabor - CS 46510, 35065 Rennes CEDEX\\
          prenom.nom@irisa.fr,\\
          \http{http://www-druid.irisa.fr}\\

 }

\resume{
Le \emph{crowdsourcing} se définit par l'externalisation de tâches à une foule de contributeurs. 
La foule est très diversifiée sur ces plateformes et inclut des contributeurs malveillants attirés par la rémunération des tâches et ne les réalisant pas consciencieusement. 
Il est essentiel d'identifier ces contributeurs afin de ne pas considérer leurs réponses. 
Tous les contributeurs n'ayant pas les mêmes aptitudes pour une tâche, il semble pertinent d'accorder un poids à leurs réponses en fonction de leurs qualifications. 
Ce papier, publié lors de la conférence ICTAI 2019, propose une méthode, MONITOR, pour l'estimation du profil du contributeur et l'agrégation des réponses par la théorie des fonctions de croyance.
}

\summary{
Crowdsourcing is characterized by the externalization of tasks to a crowd of workers. 
The crowd is very diversified due to tasks simplicity, but the payment attract malicious workers. 
It is essential to identify these malicious workers in order not to consider their answers. 
In addition, not all workers have the same qualification for a task, so it might be interesting to give more weight to those with more qualifications. 
This paper, published during the ICTAI 2019 conference, proposes a method for characterizing the profile of contributors and aggregating answers using the theory of belief functions to estimate uncertain and imprecise answers. 
}

\begin{document}
%
\section{Introduction}

Le terme \emph{crowdsourcing} définit l'externalisation de tâches à une foule de contributeurs.
Dans les plateformes de \emph{crowdsourcing}, les contributeurs réalisent des micro-tâches pour une faible rémunération. 
%
L'étude démographique de~\cite{kazai12} montre que la foule dans ces plateformes est importante et diversifiée, introduisant une diversité dans les profils des contributeurs. 
Généralement, ceux-ci sont attentifs à la tâche, mais il existe néanmoins des contributeurs attirés par l'aspect financier, répondant aléatoirement et rapidement que nous nommerons \emph{spammers}.
Compte tenu de cette diversité de profils, il est difficile d'obtenir des données fiables et pertinentes issues des plateformes de \emph{crowdsourcing}. 
%
De plus, les tâches confiées sont souvent des questionnaires à choix multiples pour lesquels 
le contributeur doit renseigner une réponse précise parmi l'ensemble de possibilités proposées. 
Or, cette interface est contraignante, car en cas d'hésitation entre plusieurs réponses, le contributeur doit
en choisir une, ce qui peut l'amener à répondre aléatoirement, perdre confiance en ses capacités et générer des données peu fiables. 

La méthode communément utilisée dans les plateformes de \emph{crowdsourcing} pour l'agrégation des réponses est le Vote Majoritaire (MV) qui consiste à sélectionner la réponse renseignée par la majorité des contributeurs. 
Cependant, cette méthode ne considère pas l'incertitude possible sur la réponse et n'est donc pas robuste face aux \emph{spammers}. 
Une alternative est l'utilisation de l'algorithme Expectation-Maximisation (EM) comme le font~\cite{raykar12} afin d'estimer la sensibilité et la spécificité des contributeurs, et agréger les réponses. 
Dans cette étude, les auteurs s'accordent sur le fait que EM est plus performant que le MV, 
néanmoins, ils ne considèrent pas le comportement du contributeur dans l'estimation de son profil. 
Si un \emph{spammer} peut parfois donner des réponses correctes, son comportement dans la réalisation de la tâche le différencie d'un contributeur sérieux.
Enfin, les méthodes abordées ci-dessus ne permettent pas de considérer l'imprécision éventuelle du contributeur. 

Ce papier est une traduction de nos travaux\footnote{Ces travaux sont partiellement financés par le Conseil Départemental des Côtes d'Armor.},~\cite{thierry19}, qui répondent aux problématiques énoncées. 
Pour ce faire, notre méthode MONITOR estime le profil du contributeur en prenant en considération sa qualification et son comportement. 
Une modélisation et une agrégation des contributions par la théorie des fonctions de croyance est également abordée.
Le reste du papier est organisé comme suit, la section 2 présente la théorie des fonctions de croyance, la section 3 synthétise l'état de l'art. 
MONITOR est introduite dans la section 4 et les expérimentations menées sont décrites dans la section 5. 
La section 6 conclut ce papier. 

\section{Théorie des fonctions de croyance}

La théorie des fonctions de croyance 
modélise l'incertitude et l'imprécision de sources imparfaites. 
Soit le cadre de discernement $\Omega=\{\omega_0, \ldots ,\omega_n\}$ défini comme un ensemble de classes $\omega_i$ exclusives et exhaustives. 
Dans le cadre du \emph{crowdsourcing}, un contributeur $c$ est une source d'information, et $\Omega$ est composé des réponses proposées à une question $q$. 
Une fonction de masse est définie pour $c$ à $q$ par $m_{cq}^\Omega:2^\Omega \rightarrow [0,1]$, telle que~: 
$\sum_{X \in 2^{\Omega}} m_{cq}^\Omega(X) = 1$.

Soit $X \in 2^\Omega$, la masse $m_{cq}^\Omega(X)$ caractérise la croyance du contributeur $c$ en la réponse $X$ à la question $q$.
Si $m_{cq}^\Omega(X)>0$ alors $X$ est appelé élément focal. 
L'élément vide symbolise l'ouverture sur le monde, dans le cas de fonctions de masse normalisées $m_{cq}^\Omega(\emptyset)=0$.  
$\Omega \in 2^\Omega$ symbolise l'ignorance, si $m_{cq}^\Omega(\Omega)=1$ alors le contributeur ignore totalement quelle est la bonne réponse. 
Généralement, une fonction $m_{cq}^\Omega(X)=1$, $X \in 2^\Omega$, est appelée fonction de masse catégorique. 
Le contributeur est alors absolument certain de sa réponse qui peut être imprécise. 
Une autre fonction spécifique est la fonction de masse à support simple $(X^w)$~:
\begin{equation} 
    \left \{
    \begin{array}{l}
        m_{cq}^\Omega(X) = w   \mbox{ avec }  X \in 2^{\Omega} \setminus \Omega\\
        m_{cq}^\Omega(\Omega) = 1 - w 
    \end{array}
    \right.
    \label{eq:m_simple}
\end{equation}
La fonction~\eqref{eq:m_simple} traduit le fait que $c$ croit en $X$ mais pas totalement.
En cas de doute sur la fiabilité d'une source $c$, un coefficient d'affaiblissement $\alpha \in [0,1]$ peut modéliser sa fiabilité~: 
\begin{equation}
    \begin{array}{l}
        m_{cq}^{\Omega,\alpha}(X) = \alpha m_{cq}^\Omega(X), \forall X \in 2^{\Omega} \setminus \Omega\\
        m_{cq}^{\Omega,\alpha}(\Omega) = 1 - \alpha(1 - m_{cq}^{\Omega}(\Omega))
    \end{array}
\end{equation}
Si $c$ n'est absolument pas fiable alors $\alpha=0$ et toute la masse est affectée à $\Omega$. 
L'affaiblissement permet de réduire les conflits qui surviennent lors de la combinaison.

De nombreux opérateurs de combinaison existent pour la fusion des informations {\em via} la théorie des fonctions de croyance, le plus courant est l'opérateur de combinaison conjonctive~: $m_{Conj}^\Omega$. 
Cet opérateur nécessite que les sources soient fiables, distinctes et indépendantes.
L'opérateur de combinaison conjonctive diminue l'imprécision sur les éléments focaux et accroît la croyance des éléments concordants entre les différentes sources d'information $c$. 
La masse $m_{Conj}^\Omega(\emptyset)$ représente le conflit global de la combinaison, elle est non nulle lorsque les sources sont en conflit. 
Afin de rester en monde clos,~\cite{yager87} interprète le conflit global comme une ignorance totale et propose pour $X \in 2^{\Omega}$ l'opérateur suivant~:
 \begin{equation} 
  \begin{array}{l}
    \displaystyle m_{Y}^\Omega(X)= m_{Conj}^\Omega(X), X \neq \emptyset, X \neq \Omega \\
    m_{Y}^\Omega(\Omega) = m_{Conj}^\Omega(\Omega) + m_{Conj}^\Omega(\emptyset) \\
    \label{eq:mYconj}
  \end{array}
\end{equation}
La combinaison se fait toujours sur le même cadre de discernement. 
Si une combinaison de sources d'information considérant des cadres de discernement différents est souhaitée, une extension vide peut être réalisée avant la combinaison~:
\begin{eqnarray} 
    m^{\Omega \uparrow \Omega \times \Theta}(B) =
    \left \{
    \begin{array}{l}
        m^\Omega(A) \mbox{ si } B = A \times \Theta, \forall A\subset \Omega \\
        0 \mbox{ sinon }
    \end{array}
    \right.
    \label{extension}
\end{eqnarray}
Afin de prendre une décision sur les éléments de $\Omega$, la probabilité pignistique est utilisée~:
\begin{equation}
 \displaystyle betP(X) = \sum_{Y \in 2^{\Omega}, Y \neq \emptyset} \frac{|X \cap Y|}{|Y|} \frac{m^\Omega(Y)}{1 - m^\Omega(\emptyset)}
 \label{eq: betP}
\end{equation}
L'élément $\omega_i \in \Omega$ pour lequel $betP(\omega_i) = \max_{\omega \in \Omega}betP(\omega)$ est sélectionné.
%
Les travaux utilisant cette théorie sont présentés dans la section suivante.

\section{État de l'art}

Deux approches sont différenciées ici, l'une utilise des données d'or et considère des réponses précises, l'autre n'utilise pas ces données et permet au contributeur d'être imprécis.
\cite{dubois19} utilisent des données d'or pour construire un graphe de référence orienté. 
Un autre graphe est réalisé à partir des contributions et est ensuite comparé au graphe de référence par la théorie des fonctions de croyance. 
La distance calculée entre les graphes permet d'estimer l'expertise du contributeur. 
La méthode de~\cite{abassi18} utilise également des données d'or, mais aussi le MV et une distance logarithmique afin d'estimer le profil du contributeur. 
Des fonctions de masse sont associées aux réponses des contributeurs d'après leur profil~: catégorique pour l'``Expert'', à support simple pour le ``Bon'' contributeur et l'ignorance pour le ``Mauvais''. 
Malheureusement, il n'est pas toujours possible d'avoir des données d'or, notamment pour les campagnes de \emph{crowdsourcing} requérant un avis subjectif de la part du contributeur.
Les approches développées ci-dessous s'affranchissent de ces données.

Pour la méthode cascade de~\cite{koulougli16}, un test de qualification est réalisé par le contributeur avant l'exécution de la tâche afin de déterminer son profil~: ``ignorant, peu compétent, moyennement compétent, compétent, expert''. 
À ce profil est associé un coefficient d'affaiblissement allant de 0 (ignorant) à 1 (expert). 
Au cours de la campagne, le contributeur renseigne son degré de confiance en sa réponse.
Ce degré est utilisé pour calculer une fonction de masse modélisant sa contribution, affaiblie par le coefficient de son profil.
Ces fonctions sont ensuite combinées en cascade. 
Les auteurs montrent que leur méthode offre de meilleurs résultats pour l'agrégation des réponses devant le MV et EM.

\cite{brjab16} modélisent les contributions par des fonctions de masse et calculent les degrés d'exactitude ($IE_c$) et de précision ($IP_c$) de chaque contributeur. 
$IE_c$ mesure l'exactitude de la réponse du contributeur comparée au reste de la foule.
Pour ce faire, les auteurs font l'hypothèse que la majorité des contributeurs a raison. 
$IP_c$ mesure la dispersion des réponses du contributeur pondérée par sa croyance. 
Ces deux degrés sont ensuite utilisés pour calculer le degré global d'expertise du contributeur 
$GD_c$ afin de déterminer son profil. 
Des données générées sont utilisées par les auteurs afin de comparer leur méthode à une approche probabiliste, prouvant qu'ils obtiennent de meilleurs résultats que cette dernière. 
Bien qu'$IE_c$ soit pertinent sous l'hypothèse de~\cite{brjab16}, celle-ci n'est pas toujours vérifiée dans le cadre du \emph{crowdsourcing}. 
$IP_c$ est plus intéressant puisqu'il ne dépend pas de cette hypothèse, c'est pourquoi nous l'avons retenu dans MONITOR.

\section{MONITOR}

\begin{figure}[t]
\begin{center}
 \includegraphics[scale=0.25]{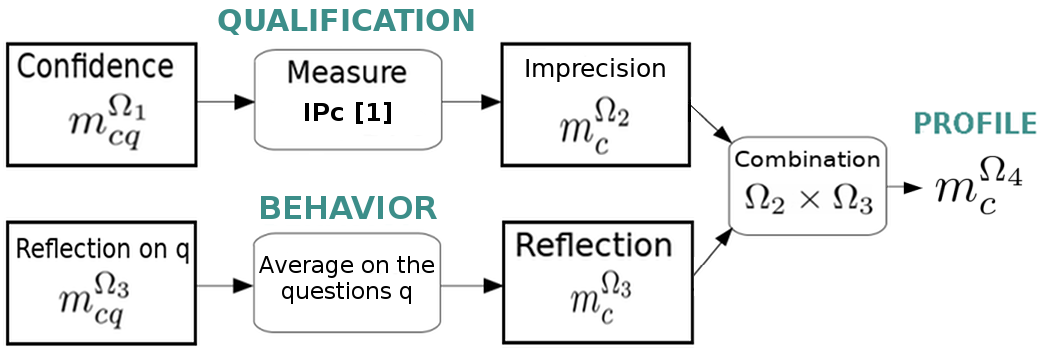}
 \caption{Schéma MONITOR.} 
 \label{fig:monitor}
\end{center}
\end{figure}

Le contributeur peut être imprécis dans sa contribution s'il hésite entre plusieurs possibilités, ainsi sa confiance en sa réponse devrait augmenter puisqu'il ne répond pas aléatoirement. 
MONITOR (illustrée par la figure~\ref{fig:monitor}) utilise la théorie des fonctions de croyance pour modéliser les réponses (\emph{Confiance}), la qualification (\emph{Imprécision}) et le comportement (\emph{Réflexion}) du contributeur afin d'établir son profil.
\paragraph{La confiance}
du contributeur $c$ en sa réponse à une question $q$ est modélisée par une fonction de masse à support simple ($X^{w_{cq}}$)~: $m_{cq}^{\Omega_1}$.
Le cadre de discernement $\Omega_{1} = \{\omega_1,\ldots,\omega_n\}$ est composé des réponses proposées. 
La masse $w_{cq}$ associée à la réponse de $c$ est une valeur numérique de la confiance qu'il a en sa contribution. 

\paragraph{L'imprécision}
du contributeur représente sa qualification pour la tâche. 
Il peut être précis ``P'' ou imprécis ``NP''. 
La fonction de croyance associée sur le cadre de discernement \linebreak $\Omega_2 = \{P, NP\}$ est donnée par l'équation~\eqref{eq:m2}, avec $\beta$ un facteur d'affaiblissement.
\begin{eqnarray} 
    \left \{
    \begin{array}{l}
        \displaystyle m_{c}^{\Omega_2}(P)= \beta*IP_c  \\
        \displaystyle m_{c}^{\Omega_2}(NP) = \beta* (1 - IP_c)\\
        m_{c}^{\Omega_2}(\Omega_2) =1 - \beta
    \end{array}
    \right.
    \label{eq:m2}
\end{eqnarray}
\paragraph{La réflexion}
du contributeur est associée à sa conscienciosité et définit ici son comportement.
En effet, le comportement est lié à la personnalité, aussi MONITOR se réfère aux \emph{Big Five}\footnote{Ouverture à l'expérience, Conscienciosité, Extraversion, Agréabilité, Névrosisme.}, pour la caractérisation des traits de personnalité. 
Cette idée est en accord avec les travaux de~\cite{kazai12} qui ont introduit les \emph{Big Five} dans un contexte de \emph{crowdsourcing} 
concluant à une forte corrélation entre la conscienciosité du contributeur et la justesse de ses résultats.
Pour estimer la réflexion du contributeur, son temps de réponse, supposé connu, est utilisé.
Plus le temps de réalisation de la tâche est long, plus le contributeur est consciencieux et réfléchi ``R''. 
À l'inverse, si un contributeur répond rapidement, de façon non réfléchie ``NR'', cela peut être dû à une contribution aléatoire ou instinctive. 
Le cadre de discernement pour la réflexion est $\Omega_3 = \{R, NR\}$. 
Soit $X \in 2^{\Omega_3}$ indiquant la réflexion de $c$ à $q$, la fonction de masse associée à $X$ est définie par 
$m_{cq}^{\Omega_3}(X) = g(T_{cq},T_{0q},X)$, 
où $T_{cq}$ est le temps de réponse de $c$ à $q$ et $T_{0q}$ le temps théorique attendu à $q$. 
La fonction $g$ est définie par~\cite{thierry19}. 
Une fois la masse $m_{cq}^{\Omega_3}$ calculée pour chaque question, une moyenne sur $q$ est effectuée afin d'obtenir la masse $m_{c}^{\Omega_3}$ qui modélise la réflexion globale du contributeur.

\paragraph{Le profil}
du contributeur est estimé par le produit cartésien des cadres de discernement $\Omega_4 = \Omega_2 \times \Omega_3$. 
Pour obtenir $\Omega_4$ les fonctions $m_c^{\Omega_2 \uparrow \Omega_4}$ et $m_c^{\Omega_3 \uparrow \Omega_4}$ sont calculées, puis l'opérateur conjonctif de Yager, leur est appliqué. 
Afin de prendre une décision, $m_c^{\Omega_4}$ est transformée en probabilité pignistique. 
Le profil présentant la probabilité la plus élevée est assigné au contributeur. 
Les profils définis sur $\Omega_4$ sont les suivants~:
\begin{itemize}
 \item {\bf L'expert \{NR,NP\}} est plus qualifié pour réaliser la tâche que la moyenne des contributeurs. 
 Ses réponses sont instinctives et il s'autorise à être imprécis en cas de doute.
 \item {\bf Le contributeur flou $\{R, NP\}$} est imprécis dans ses réponses et consciencieux. 
 \item {\bf Le contributeur catégorique $\{R, P\}$} est consciencieux mais ne prend pas l'opportunité qui s'offre à lui d'être imprécis. 
 \item {\bf Le \emph{spammer} $\{NR, P\}$} est uniquement intéressé par la rémunération de la tâche, il répond rapidement, aléatoirement et de façon précise.
\end{itemize}


\section{Données et résultats expérimentaux}



Les données utilisées pour valider MONITOR proviennent d'une campagne de \emph{crowdsourcing} consistant à noter des enregistrements sonores d'après l'échelle de qualité~: (1) mauvais (2) pauvre (3) correct (4) bon (5) excellent.
Cette échelle de qualité constitue $\Omega_1$.
Lorsque le contributeur donne sa réponse, il doit également spécifier son degré de confiance ($w_{cq}$)~: très sûr (0.99), plutôt sûr (0.75), moyennement sûr (0.5), peu sûr (0.25), pas sûr (0.01). 
Les degrés de confiance $w_{cq}$ sont échelonnés de 0 à 1 par palier de $\frac{1}{|\text{Échelle de confiance}|}$ soit ici 0.25.
Les valeurs $w_{cq}$ associées à ``très sûr'' et ``pas sûr'' sont respectivement diminuées et augmentées de 0.01 afin de ne pas avoir de fonctions de masse catégoriques.
Il est, en effet, préférable de maintenir une incertitude sur la réponse du contributeur.

La campagne est composée de 4 HITs (\emph{Human Intelligent Tasks}), chacun incluant 12 enregistrements.
La qualité sonore de 5 d'entre eux est connue.
Ces données d'or ne sont pas utilisées dans l'étape de modélisation proposée mais elles permettent de valider les résultats obtenus par MONITOR.
93 contributeurs ont réalisé la campagne.
Chacun d'entre eux a noté les 12 enregistrements des 4 HITs, ce qui fait un total de 4 464 contributions.
Parmi ces contributions, 21.6~\% sont imprécises, provenant de 70 contributeurs parmi les 93 qui composent la foule. 
Cette utilisation importante de l'imprécision démontre l'intérêt de cette forme d'expressivité correspondant à un besoin réel de la foule.

\begin{figure}[t]
\begin{center}
  (a)
 \includegraphics[scale=0.24]{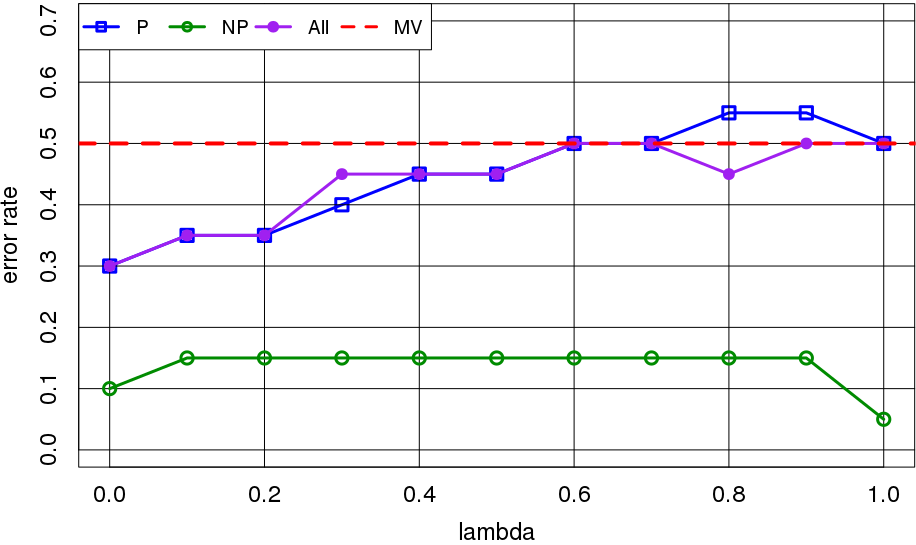}
  (b)
 \includegraphics[scale=0.24]{img/eval_reflection.png}
 \caption{Évolution du taux d'erreur des contributeurs d'après leur (a) {\bf précision}, (b) {\bf réflexion}.} 
 \label{fig:ev_imp}
\end{center}
\end{figure}
\begin{figure}[t]
\begin{center}
 \includegraphics[scale=0.3]{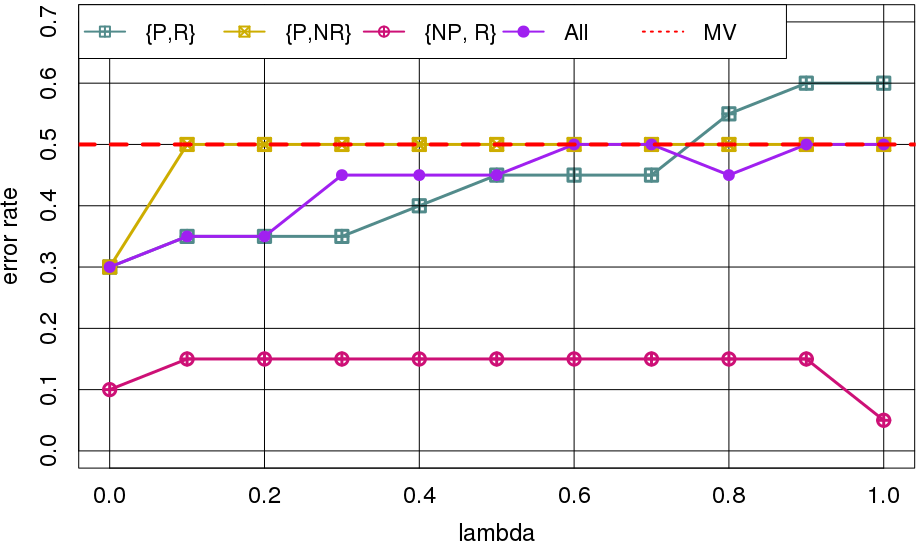}
 \caption{Évolution du taux d'erreur des contributeurs d'après leur {\bf profil}.} 
 \label{fig:ev_pro}
\end{center}
\end{figure}

\paragraph{L'agrégation des réponses} est réalisée en sommant les moyennes des réponses précises $m_{p}^{\Omega_1}$ et imprécises $m_{ip}^{\Omega_1}$ pondérées par un coefficient $\lambda$~:
$  m_{\lambda}^{\Omega_1} = \lambda*m_{p}^{\Omega_1} + (1-\lambda)*m_{ip}^{\Omega_1} $.
%
Une fois $m_{\lambda}^{\Omega_1}$ calculée pour la question $q$, cette fonction est transformée en probabilité pignistique afin de déterminer la bonne réponse.
Différentes valeurs de $\lambda$ ont été testées sur les 5 données d'or des 4 HITs.
Pour chaque $\lambda$ testé la réponse obtenue par le maximum de probabilité est comparée à la réponse attendue afin de calculer un taux d'erreur sur ces 20 questions.
L'évolution de ce taux d'erreur sur les données d'or est représenté par la courbe ``All'' des figures~\ref{fig:ev_imp}, \ref{fig:ev_pro}.
On observe sur ces figures que le taux d'erreur diminue avec $\lambda$, ainsi accroître le poids accordé aux réponses imprécises accroît la justesse des résultats.
De plus, cette méthode d'agrégation offre un taux d'erreur généralement plus faible que le MV.
%
\paragraph{Évaluation de la qualification.}
Afin d'estimer l'imprécision du contributeur, $\beta$ est fixé à 0.8 dans l'équation~\eqref{eq:m2}, et $m_{c}^{\Omega_2}$ est transformée en probabilité pignistique.
Le groupe des contributeurs imprécis n'est composé que de 1.08~\% de la foule, ce qui s'explique par le fait qu'il n'est pas commun pour les contributeurs d'être imprécis dans leurs réponses. 
Sur la figure \ref{fig:ev_imp} (a), le taux d'erreur des contributeurs imprécis ``NP'' est plus faible que celui des contributeurs précis ``P'' et celui du MV.
Pour $\lambda = 0.8$ et $0.9$, le taux d'erreur est supérieur à celui du MV, mais ce n'est pas le cas pour l'agrégation de la foule, ce qui signifie que les réponses des contributeurs imprécis ont un impact important sur l'agrégation globale.
%
En sélectionnant les contributeurs les plus susceptibles d'être imprécis le taux d'erreur obtenu est le plus faible de tous, 
l'imprécision est donc bien bénéfique dans le cadre du \emph{crowdsourcing}.
\paragraph{Évaluation du comportement.}
Le temps $T_{0q}$ est la durée (en secondes) de l'enregistrement à noter.
%
%
L'estimation par MONITOR de la réflexion du contributeur, pour les tests sur les données d'or, se fait en transformant la fonction de masse $m_{c}^{\Omega_3}$ en probabilité pignistique.
MONITOR classifie 65,6~\% des contributeurs comme réfléchis ``R'', ce qui semble correct, car il est attendu que la majorité des contributeurs soient réfléchis, les experts et \emph{spammers} étant plus rares au sein de la foule.
Pour 3,2~\% des contributeurs, il n'est pas possible de choisir entre réfléchi ou non. 
Les 31,2~\% restants ne sont pas réfléchis ``NR'' et sont donc étiquetés comme tels. 
%
Considérons l'agrégation des réponses des contributeurs d'après leur réflexion (figure~\ref{fig:ev_imp} (b)), le taux d'erreur des contributeurs ``NR'' converge rapidement vers celui de MV.
Comme ce taux d'erreur est plus élevé que pour l'ensemble de la foule, cela suggère qu'il y a davantage de \emph{spammers} que d'experts.
L'agrégation différenciée des contributeurs d'après réflexion est positive puisque les contributeurs réfléchis ont le taux d'erreur plus faible pour $\lambda < 0.8$.  
%
\paragraph{Évaluation du profil.}
MONITOR classifie 62,4~\% des contributeurs comme catégoriques \{P,R\}, 30,1~\% comme \emph{spammers} \{P,NR\}, et 1,1~\% comme flous \{NP,R\}.
Il n'y a pas de contributeurs estimés experts \{NP,NR\}, qui devraient être pour cette campagne des contributeurs ayant une oreille absolue.
Mais~\cite{takeuchi93} estiment que moins de 0,012~\% de la population possède l'oreille absolue.
Il devrait donc y avoir 0,93 contributeurs avec cette capacité présents dans la foule de cette étude, ce qui explique l'absence d'expert.
Pour 6,45~\% des contributeurs il y a une indécision sur le profil dû à l'estimation de la réflexion, le maximum de probabilité pignistique est donné pour \{P,R\}$\cup$\{P,NR\}. 
Pour $\lambda <0.8$ (figure~\ref{fig:ev_pro}), les \emph{spammers} ont le taux d'erreur le plus élevé, cette mauvaise performance s'explique par leurs réponses aléatoires.
Les contributeurs catégoriques sont aussi performants, voire meilleurs que l'ensemble de la foule et 
le contributeur flou, est le plus performant.

Globalement, le taux d'erreur augmente avec $\lambda$ pour tous les profils, donc donner plus de poids aux réponses imprécises est bénéfique pour l'agrégation des données. 
Les taux d'erreur des groupes de contributeurs d'après leurs profils sont en accord avec les résultats escomptés, ce qui renseigne une bonne estimation du profil du contributeur par MONITOR.

\section{Conclusion}

Les principales problématiques du \emph{crowdsourcing} sont l'ergonomie des tâches, l'estimation du profil des contributeurs et l'agrégation des réponses.
Ce papier est une traduction de nos travaux~\cite{thierry19} répondant à ces problématiques en offrant au contributeur la possibilité d'être imprécis dans ses réponses en cas d'hésitation. 
Pour modéliser l'incertitude et l'imprécision des contributions, la théorie des fonctions de croyance est utilisée. 
Les tests effectués sur des données réelles pour la validation de MONITOR montrent une estimation du profil très positive, de plus la méthode d'agrégation des contributions proposée est plus efficace que le MV.
Malheureusement, les données utilisées ne permettent pas d'observer tous les types de profils définis en raison du type d'expertise requis par la campagne de \emph{crowdsourcing}. 
Dans nos futurs travaux, nous mènerons de nouvelles campagnes nécessitant des compétences moins spécifiques de la part des contributeurs et comparerons MONITOR à EM pour l'agrégation des réponses.
Nos perspectives visent également à intégrer d'autres traits de personnalité pour améliorer l'estimation du comportement du contributeur.

\bibliographystyle{rnti}
\bibliography{references.bib}



\end{document}